\documentclass[letterpaper, 10 pt, conference]{ieeeconf}  

\IEEEoverridecommandlockouts                              

\usepackage{cite}
\usepackage{amsmath,amssymb,amsfonts}
\usepackage{algorithmic}
\usepackage{graphicx}
\usepackage{textcomp}
\usepackage{xcolor}

\usepackage{array}
\usepackage{stfloats}
\usepackage{url}
\usepackage{color}
\usepackage[ruled,linesnumbered]{algorithm2e}

\title{\LARGE \bf
Learning a General Model: Folding Clothing with Topological Dynamics
}
\author{Yiming Liu$^{1}$, Lijun Han$^{1}$, Enlin Gu$^{2}$ and Hesheng Wang$^{1}$, \emph{Senior Member, IEEE}
\thanks{*This work was supported in part by the Natural Science Foundation of China under Grant 62225309, 62073222, U21A20480 and 62361166632, in part by the Open Research Projects of Zhejiang Lab under Grant 2022NB0AB01.}
\thanks{$^{1}$ Y. Liu, L. Han, and H. Wang are with the Department of Automation, Shanghai Jiao Tong University, Shanghai 200240, China (e-mail: lijun\_han@sjtu.edu.cn; lym5634@sjtu.edu.cn; wanghesheng@sjtu.edu.cn).}%
\thanks{$^{2}$ Enlin Gu is with the School of Engineering, University of Pennsylvania, PA, 19104-6303, USA (e-mail: guenlin@seas.upenn.edu).
}}

\begin{document}
\maketitle
\thispagestyle{empty}
\pagestyle{empty}

\begin{abstract}
The high degrees of freedom and complex structure of garments present significant challenges for clothing manipulation. In this paper, we propose a general topological dynamics model to fold complex clothing.
By utilizing the visible folding structure as the topological skeleton, we design a novel topological graph to represent the clothing state. 
This topological graph is low-dimensional and applied for complex clothing in various folding states. It indicates the constraints of clothing and enables predictions regarding clothing movement.
To extract graphs from self-occlusion, we apply semantic segmentation to analyze the occlusion relationships and decompose the clothing structure. The decomposed structure is then combined with keypoint detection to generate the topological graph. 
To analyze the behavior of the topological graph, we employ an improved Graph Neural Network (GNN)  to learn the general dynamics.
The GNN model can predict the deformation of clothing and is employed to calculate the deformation Jacobi matrix for control.
Experiments using jackets validate the algorithm's effectiveness to recognize and fold complex clothing with self-occlusion.
\end{abstract}

\section{Introduction}
Clothing manipulation is an intricate household task. Equipping robots with the capability to manipulate soft fabrics will enhance their intelligence and improve the quality of people's lives. 
However, the inherent complexity of clothing presents challenges. As a pliable object, fabric possesses infinite degrees of freedom and often undergoes significant deformation with self-occlusion. Early approach \cite{liuFastSimulationMassspring2013} utilizes mesh-based model to describe the clothing state, whose efficiency is limited by complex dynamics. Furthermore, unlike simpler fabrics such as towels, clothing is not a flat 2D structure. Even when fully unfolded, it still retains Local overlap, further complicating the analysis of its behavior. 
Although recent research has made progress in tasks such as hanging \cite{DavidTowardsClothesHanging2021}, unfolding \cite{FuFlingFlowLLMDrivenDynamic2024}, and folding \cite{LonghiniAdaFoldAdaptingFolding2024}, most methods are limited to simple cloth like towels or T-shirts. The geometries of these clothes are relatively rectangular and approximated as one-layer. Methods for handling complex clothing with long sleeves and other attached parts are still under development.
\par
For developing clothing manipulation algorithms, it is essential to understand how humans manipulate clothing. Instead of modeling the garment's deformation precisely, we extract key features by analyzing the garment's structure. By predicting the behavior of those topological features, we can develop a manipulation policy for clothing deformation tasks.
This topology-based approach, which includes perception, modeling, and control of soft objects, has already been applied in the manipulation of deformable linear objects\cite{zhouReactiveHumanRobot2024}. However, extracting topological features from clothing is more challenging, limiting further application.
\par

\begin{figure}[!t]
        \centering
        \includegraphics[width=0.48\textwidth,trim=5 1 5 1,clip]{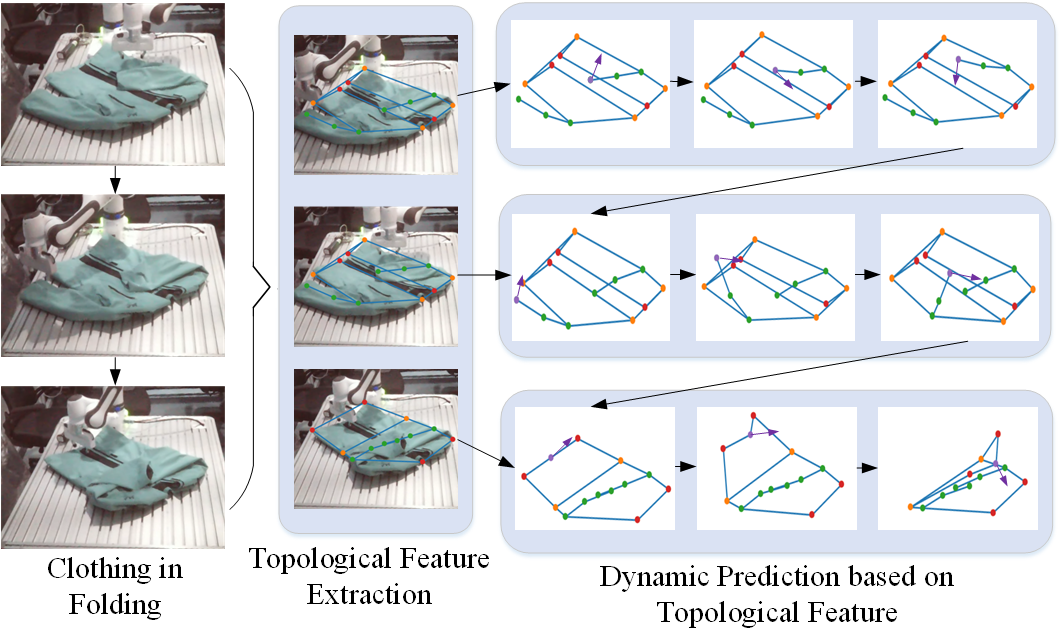}
        \caption{We propose a new topology-based representation of clothing states. By characterizing clothing occlusions and keypoints, this representation can be extracted and generated from complex clothing in different folding states. A general GNN is applied to obtain the dynamics model of the topological graph.}
        \label{main_struct}
\end{figure}

In this paper, we propose a topology-based method for clothing folding. By characterizing the topological skeleton of clothing, we can develop a topological representation for clothing folding (see Fig. \ref{main_struct}). This representation reduces the feature dimension and adapts to complex garments in various folding states.
Specifically, by combining keypoint detection with occlusive layer semantic segmentation, we present the process for generating topological graphs from different folding states and under self-occlusion conditions.
Considering the sparsity of the 2D topology skeleton, we use Graph Neural Network (GNN) and improve the message propagation mechanism to learn the dynamics of clothing. An additional aggregation and update mechanism for nodes within the same region is introduced to enhance network efficiency and model the general dynamics of topological graphs.
Based on the extraction of topological graphs and their general dynamics, a Jacobian-based controller is designed to execute the folding task in stages. 
\par
In summary, the contributions of this paper are as follows:
\begin{itemize}
        \item We define a topological graph that can represent complex clothing in various folding states. By combining keypoint detection and occlusive layer semantic segmentation, this graph can be extracted from self-occluded garments.
        \item We propose a general dynamics model for the topological graph to predict the deformation of clothing. This model enables the Jacobi-based control, which can be applied to fold complex clothing.
        \item We developed the platform and conducted the experiments on folding jackets. The results validate the effectiveness of our perceiving and controlling method.
\end{itemize}

\section{Related Work}
\textbf{Clothing Feature Perception:}
The infinite dimensionality and self-occlusion of clothing present significant challenges for clothing feature perception. 
Traditionally, geometric features, such as corner points \cite{LiFoldingDeformableObjects2015} and clothing seams \cite{KondoDevelopmentand2022}, have been employed in state perception. The application of these geometric features is limited by the lack of relationships between different features, making clothing modeling challenging. 
Leveraging the DeepFashion dataset\cite{LiuDeepFashionPoweringRobust2016}, approaches such as \cite{WangAttentive2018} and \cite{GustavssonClothManipulationBased2022} detect the corresponding semantic keypoints and construct topological skeletons. However, under self-occlusion, the difficulty to identify clothing categories and the lack of the keypoints limit its application. 
Recently, the normalized canonical space is proposed to reconstruct and track the clothing meshes from point clouds \cite{chigarmentnetscategorylevelpose2021}, \cite{xuegarmenttrackingcategorylevelgarment2023}. These methods still rely on the priors of clothing categories and are only trained in simulation.
\par
In this paper, we propose a novel algorithm for keypoint detection and topological relationship generation that is independent of clothing categories. 

\textbf{Cloth Manipulation:}
To achieve the desired cloth configuration, cloth manipulation research can be classified into two categories: model-free methods and model-based methods.
Model-free methods employ reinforcement learning \cite{WuLearningto2020}, \cite{HietalaLearning2022} or imitation learning \cite{GanapathiLearning2021} to learn the mapping functions from perception to action. However, These algorithms are tailored to specific tasks and struggle to generalize across different types of fabrics and operations. 
On the other hand, Model-based methods represent states using particles \cite{MacklinUnifiedParticlePhysics2014} or images \cite{LippiLatent2020}, employing analysis techniques such as MSD \cite{todorovmujocophysicsengine2012} and PBD \cite{MacklinUnifiedParticlePhysics2014}, or network architectures like LSTM \cite{hoquevisuospatialforesightphysical2022} and GNN \cite{libaomeshgraphnetrpimprovinggeneralization2023} to learn state transition relationships. Through optimization control, these methods can generate the optimal control policy. However, due to the high dimensionality of state representation, model-based methods require significant computational resources, limiting their applicability to fabrics with complex structures.
\par
In this paper, we propose a new model-based clothing folding method that leverages topological graphs with low dimension.


\section{Proposed Method} 
We study the task of folding complex garments. Initially, the clothing is allowed to be unfolded entirely while not being flattened. It exposes the overall structure but still exhibits wrinkles and may be self obstructed, with the sleeves' position remaining uncertain. 
These conditions impose challenges on the detection of keypoints and the accurate matching of the clothing outline template. 
\par
\begin{figure*}[htbp]\centering 
        \includegraphics[width=0.95\textwidth,trim=1 1 1 1,clip]{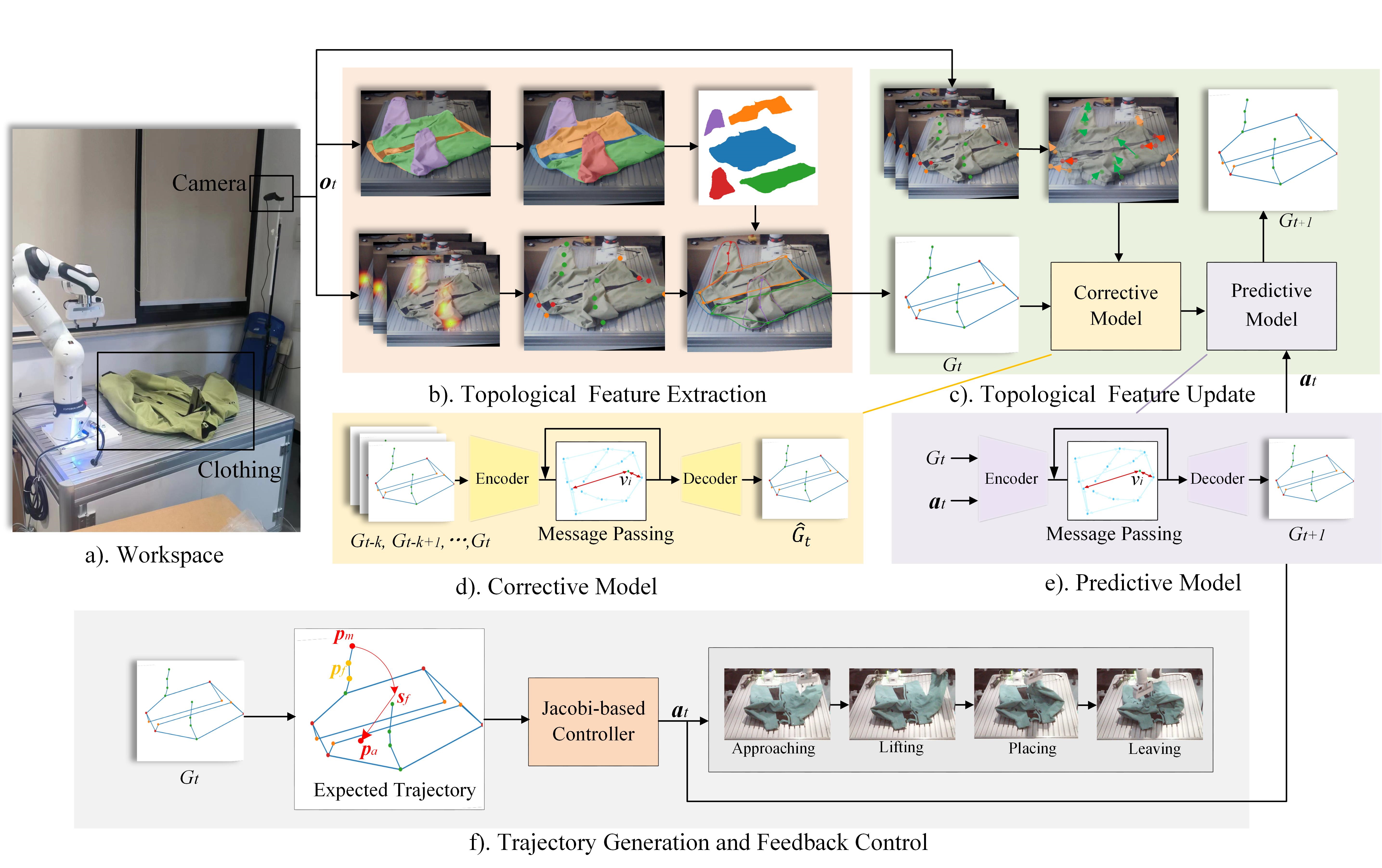}
        \caption{Overview. (a) In the workspace, an RGB image observation $\boldsymbol{o}_t$ is captured by the camera. (b) Semantic keypoints and occlusive layer semantic segmentation are extracted from the observation and matched to construct the topological graph, $G_t$. (c) During manipulation, the tracked keypoints are utilized to refine $G_t$ via a GNN-based corrective model, while the new state, $G_{t+1}$ is predicted by a GNN-based predictive model. The corrective model (d) and the predictive model (e) employ the message passing mechanism to fit the expected process.  The DJM is obtained from the predictive model by back propagation. (f) Based on the topological state, the folding trajectory is generated in 4 stages: approaching, lifting, placing, and leaving. The robot actions, $\boldsymbol{a}_t$ are calculated by the Jacobian controller to follow the trajectory.
        }
        \label{pipline}
\end{figure*}
At time $t$, we obtain a visual observation of the clothing, $\boldsymbol{o}_t\in \mathbb{R}^{W\times H \times C}$, from a fixed and calibrated camera. The robot's actions, $\boldsymbol{a}_t=(\boldsymbol{p}_{robot}, s_{gripper})$, involve the robot's end-effector position $\boldsymbol{p}_{robot}\in \mathbb{R}^3$ and the gripper status $s_{gripper}\in \{0,1\}$. The topological graph, denoted as $G_t$, is used to represent the state of the clothing. We formulate the problem as extracting $G_t$ from the observation $\boldsymbol{o}_t$ and determining the next action $\boldsymbol{a}_{t}$ based on the current clothing state $G_t$. This relationship is formalized as the functions $\mathcal{G}_0$, $\mathcal{G}$, and $\mathcal{\pi}$ in (\ref{problem}):
\begin{align}
        \begin{split}
                & G_0 = \mathcal{G}_0(\boldsymbol{o}_0) \\
                & G_{t+1} = \mathcal{G} (G_t, \boldsymbol{a}_t, \boldsymbol{o}_t+1) \\
                & \boldsymbol{a}_t = \mathcal{\pi}(G_t) \\
        \end{split}        
        \label{problem}
\end{align}
A trial is a \textit{success} when the area of clothing is smaller than a specified threshold after a given number of folds.
\par
As shown in Fig. \ref{pipline}, we define the topological graph to represent the clothing (Sec. III-A). The topological graph is extracted from the camera (Sec. III-B) and updated by the GNN-based dynamics model (Sec. III-C). With the deformation Jacobi matrix obtained from the model, the clothing folding task is implemented by Jacobi-based feedback control (Sec. III-D).

\subsection{General Topological Graph for Clothing} 
To overcome the impact of self-occlusion and variations in visible clothing structure, we introduce a heterogeneous graph $\langle V, E \rangle$ to represent the folding state of clothing. 
The clothing is segmented into stitched regions based on the creases on it. The vertex nodes $N_0$ of each region are selected to describe their shape, while the endpoints $N_1$ of the creases serve as nodes to capture the relationship between different regions.
Considering the perception and operation requirement, we only focus on the outermost creases and the corresponding regions. 
\par
Our method considers the slender sleeves as independent linear structures to handle first. The sleeve nodes $N_2$, which are sampled from the sleeves' principle axes, are employed to represent the sleeves' states. This approach simplifies the clothing shape and helps to address obstructions.
\par
In graph $\langle V, E \rangle$, nodes $V=\{\boldsymbol{v}_i\},(i=1,...,N)$ represent topological nodes. Each node, $\boldsymbol{v}_i = (\boldsymbol{p}_i, \dot{\boldsymbol{p}}_i, h_i, k_i, \boldsymbol{p}_{ci})$, involves the current position $\boldsymbol{p}_i\in \mathbb{R}^3$, velocity $\dot{\boldsymbol{p}}_i\in \mathbb{R}^3$, the distance $h_i \in \mathbb{R}$ from the plane, node type $k_i\in\{N_0,N_1,N_2\}$, and the center $\boldsymbol{p}_{ci}$ of the region the node belongs to. For crease nodes belonging to multiple regions, $\boldsymbol{p}_{ci}$ is calculated as the means of centers. Adjacent sleeve nodes are considered as an independent fabric region to calculate corresponding $\boldsymbol{p}_{ci}$. The node sets of these regions are denoted as $\Lambda_k$.
\par
To construct the graph edges $E$, nodes are connected by the contours of flat regions. Each edge $\boldsymbol{e}_{ij}$ is defined as:
\begin{align}
\boldsymbol{e}_{ij} = (|| \boldsymbol{p}_i - \boldsymbol{p}_j||, \frac{||\boldsymbol{p}_i-\boldsymbol{p}_j||}{d_{ij}}, \boldsymbol{p}_{ij}, \boldsymbol{r}_{ij})        
\end{align}
where $d_{ij}$ denotes the maximum distance between nodes $\boldsymbol{v}_i$ and $\boldsymbol{v}_j$ during deformation. The term $\boldsymbol{p}_{ij}$ refers to the farthest position of the fabric outline from the edge $\boldsymbol{e}_{ij}$, while $\boldsymbol{r}_{ij}$ indicates the direction of $\boldsymbol{p}_{ij}$ relative to $\boldsymbol{e}_i$. Together, $\boldsymbol{p}_{ij}$ and $\boldsymbol{r}_{ij}$ are employed to describe the deformation.
\par
Notably, edge features such as $d_{ij}$ and $\boldsymbol{p}_{ij}$ are difficult to directly perceive from the observation. These edge attributes are refined through the subsequent corrective model.

\subsection{Topological Graph Extraction} 
\begin{figure}[!t]\centering
    \includegraphics[width=0.48\textwidth,trim=5 1 5 15,clip]{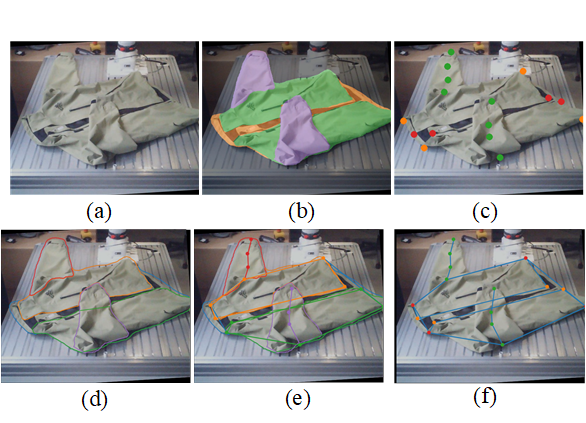}
    \caption{The process to construct the topological graph. (a) The observation. (b) The keypoints. (c) The segmentation of occlusive regions. (d) The contours of connected regions. (e) The contours matched with keypoints. (f) The result.}
    \label{feature_extraction_process}
\end{figure}

To construct the topological graph, we combine keypoint detection and semantic segmentation of occlusive relationships. 
In keypoint detection,  we use ViTPose \cite{YufeiViTPoseSimpleVision2022} to learn the keypoint heatmap $\boldsymbol{h}_k \in \mathbb{R}^{W\times H}, k\in\{0,1,2\}$. Considering that the number of nodes are uncertain in variation clothing structure, all nodes of the same type are detected by one heatmap.
To extract multiple keypoints from a heatmap, we locate the maximum value $h_{max}$ and its corresponding position $\boldsymbol{p}_{max}$ in $\boldsymbol{h}_k$. When $h_{max}$ exceeds the threshold, $\boldsymbol{p}_{max}$ is appended into the result and the heatmap is updated by adding mask $\boldsymbol{m}_k\in \mathbb{R}^{W\times H}$ to eliminate $\boldsymbol{p}_{max}$. The $(i, j)$ entry of $\boldsymbol{m}_k$, ${m}_{k,ij}$ is defined as:
\begin{align}
        \boldsymbol{m}_{k,ij} = h_{max}e^{-k_h||\boldsymbol{p}_{max}-\boldsymbol{p}_{ij} ||}
\end{align}        
where $k_h$ is a positive coefficient. $\boldsymbol{p}_{ij}$ is the position of ${m}_{k,ij}$ in the mask. This process is repeated until no keypoints can be detected. 
In semantic segmentation of the occlusive layer, $\boldsymbol{o}_t$ is segmented into four distinct regions: sleeves with no occlusion, clothing body with no occlusion, clothing body with one layer occlusion, and clothing body with two layers occlusion. The adjacent fabric regions in $\boldsymbol{o}_t$ are classified into different types. Mask2Former \cite{cheng2021maskformer} is employed for semantic segmentation. The results of keypoint detection and semantic segmentation are presented in Fig. \ref{feature_extraction_process} (b), (c).
\par
After retrieving the segmented regions, we further analyze the occlusion relationships of the clothing to identify its parts. There are two primary cases:
In one case, two non-adjacent, occluded regions are separated by an upper fabric. If they are connected by the intersection between the upper region and the smallest convex polygon that encloses both occluded regions, the two occluded regions belong to one part with the intersection.
In the other case, two adjacent fabrics on different layers are contacted by a banding area. If there exist crease nodes near the upper fabric but away from the lower one, the two regions are connected by the creases and the higher region is extended as the hidden region of the lower one. 
These two rules allow us to obtain the whole clothing regions shown in Fig. \ref{feature_extraction_process} (d).
\par
Finally, we assign detected keypoints with these parts:
Sleeve keypoints in the same region are selected and connected by distance. Crease keypoints close to overlapping regions are employed as their crease nodes. The end of sleeves, crease nodes, and vertex keypoints are assigned to their near regions and connected along the contours. The matched result is shown in Fig. \ref{feature_extraction_process} (e). 
The node features are calculated using the camera projection model, while the edge features are initialized by the shape of the contours. Thus, we obtain the topological graph as shown in  Fig. \ref{feature_extraction_process} (f). 
Additionally, incorrect keypoint detection may result in topology extraction failure, such as an inability to identify a sufficient number of keypoints for one contour. In such cases, the next observation, $\boldsymbol{o}_{t+1}$, is utilized to re-extract the topological graph.
\par
During manipulation, we use CoTracker3 \cite{NikitaCoTracker3Simplerand2024} to track the visible nodes in real-time. The position closest to the current state, which can be projected onto the same pixel in the image, is used to update the node state.

\subsection{GNN-based Dynamics Models} 
Previous research \cite{AlvaroLearningtoSimulate2020} has shown the performance of the message passing based GNN for dynamics learning. Similarly, we use GNNs to model the forward dynamics and feedback update of the topological graphs.

Define the GNN-based predictive model as $f_{forward}$ and the GNN-based corrective model as $f_{feedback}$. The two networks are composed of \textit{Encoder}, \textit{Processor}, and \textit{Decoder}. The two nets have different input and output widths in encoders and decoders. The functions of the three modules are as follows.
\par
\textit{Encoder}: The encoder module transforms the edge and node features into the latent space, $\langle V', E' \rangle$. The different node types are encoded as the corresponding type embedding vectors. For $\boldsymbol{a}_t$, the gripping status for each node is encoded as the gripper embedding vector. If $s_{gripper}=1$, the robot's position $\boldsymbol{p}_{robot}$ is concatenated to the feature of the picked node, while the features of other nodes are padded to the same length with $0$. In $f_{forward}$, the current topological graph $G_t$ with the actions $\boldsymbol{a}_t$ is encoded, while $f_{feedback}$ uses the sequence $\{G_{t-i}, \boldsymbol{a}_{t-i}\}$ as inputs. The encoder for node feature and edge feature, $f_{\varphi_E}$ and $f_{\varphi_V}$, are formulated as:
\begin{align}
        \begin{split}
                 ^0\boldsymbol{v}'_i = &f_{\varphi_V}(\boldsymbol{v}_{i|t}, \boldsymbol{v}_{i|t-1},..., \boldsymbol{v}_{i|t-l+1},\\
                                    & g(\boldsymbol{a}_t), g(\boldsymbol{a}_{t-1}),..., g(\boldsymbol{a}_{t-l+1}), \boldsymbol{s}_{em|k_i} )\\
                 ^0\boldsymbol{e}'_{ij} = &f_{\varphi_E}(\boldsymbol{e}_{ij|t}, \boldsymbol{e}_{ij|t-1},..., \boldsymbol{e}_{ij|t-l+1})\\
                 g(\boldsymbol{a}_t) = & \left\{\begin{matrix}
                        (\boldsymbol{p}_{robot},\boldsymbol{s}_{em|g}); \text{ if } s_{gripper} =1 \phantom{123}  \\
                             \phantom{12345678910}            \text{ and } ||\boldsymbol{p}_{robot}-\boldsymbol{p}_i|| < d_p\\
                        (\boldsymbol{0},\boldsymbol{s}_{em|f});  \phantom{1234}    \text{o.w.}  \phantom{12345678910}
                 \end{matrix}\right.
        \end{split}
\end{align}
where $^0\boldsymbol{v}'_i$ and $^0\boldsymbol{e}'_{ij}$ are the initial node and edge features in the latent space. $l$ is the input sequence length. $\boldsymbol{s}_{em|k_i},\boldsymbol{s}_{em|p},\boldsymbol{s}_{em|f}$ are embedding vectors for nodal types. $d_p$ is the threshold for gripping.
\par
\begin{figure}[!t]\centering
        \includegraphics[width=0.48\textwidth,trim=5 1 5 1,clip]{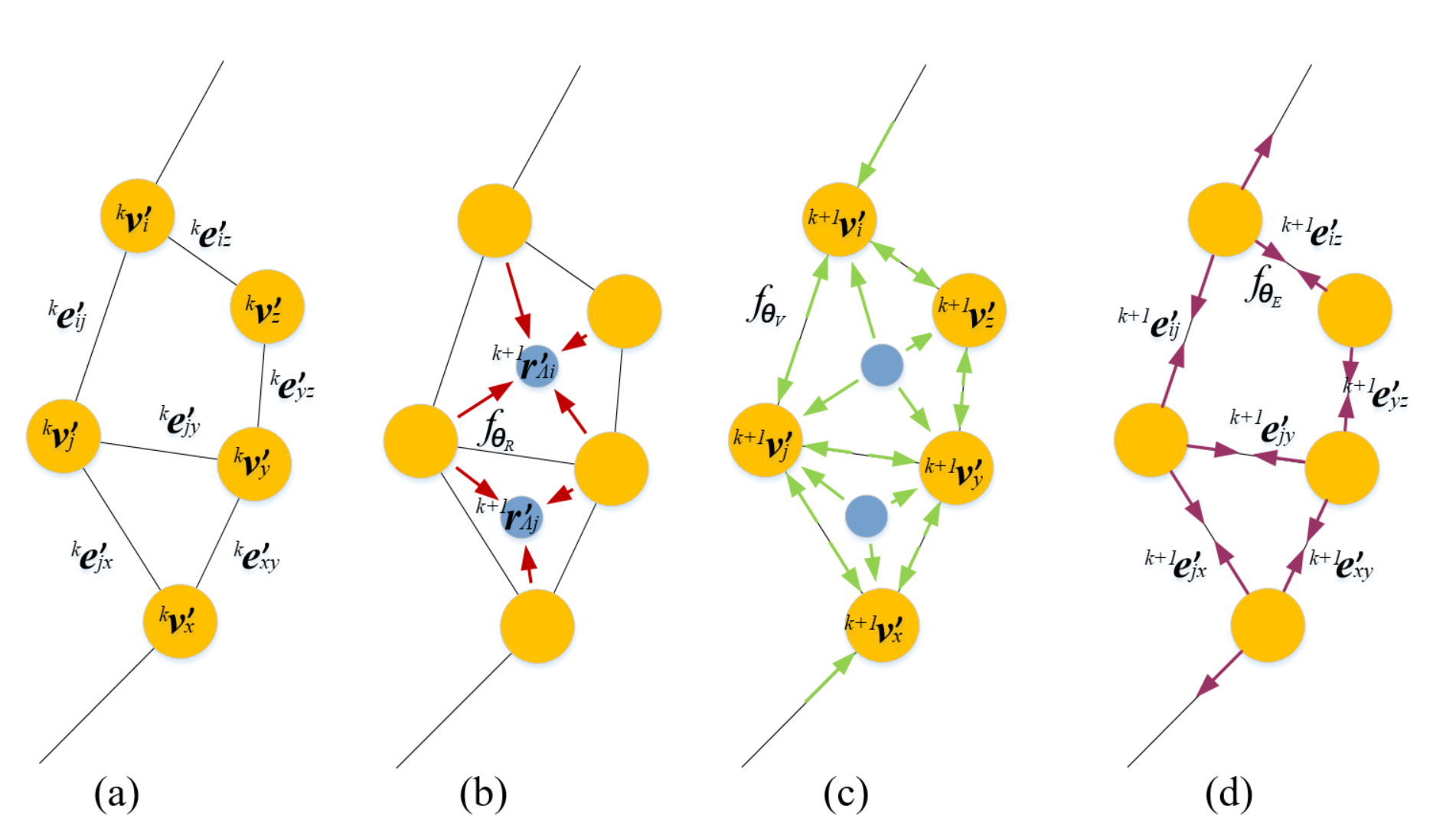}
        \caption{The process of one-step message passing in our graph net. (a) is the latent graph state to update. (b) is implemented first to obtain the regional feature $^{k+1}\boldsymbol{r}'_{\Lambda_i}$. (c) and (d) are calculated together then to update the node and edge states.}
        \label{message_passing}
    \end{figure}

\textit{Processor}: The processor is employed to calculate the new edge and node features in latent space. 
According to the message passing mechanism, features of adjacent edges and nodes are aggregated and updated iteratively. 
Besides, nodes belonging to the same region $\Lambda_i$ have strong constraints even if they are not adjacent. Therefore, we aggregate the node features as $^{k}\boldsymbol{r}'_{\Lambda_i}$ in the same region before each propagation. Then, the aggregated features  $^{k}\boldsymbol{r}'_{\Lambda_i}$ are employed to 
concatenate as a part of the adjacent nodes' features, which represents the effect of overall fabric movement.
 The process is depicted in Fig. \ref{message_passing} and formalized as follows:
\begin{align}
        \begin{split}
                ^{k+1}\boldsymbol{e}'_{ij} = &f_{\theta_E}(^k \boldsymbol{v}'_i, ^k \boldsymbol{e}'_{ij}, ^k \boldsymbol{v}'_j)\\
                ^{k+1}\boldsymbol{r}'_{\Lambda_i} = &f_{\theta_R}(\sum_{j \in \Lambda_i} {}^k \boldsymbol{v}'_j)\\
                ^{k+1}\boldsymbol{v}'_i = &f_{\theta_V}(^k \boldsymbol{v}'_{i}, \sum_{j} {}^k \boldsymbol{e}'_{ij}, \frac{1}{l_i}\sum_{j,\text{ } i \in \Lambda_j} {}^{k+1}\boldsymbol{r}'_{\Lambda_j})
        \end{split}
        \label{m_p_eq}
\end{align}
In  (\ref{m_p_eq}), $^k\boldsymbol{v}'_i$ and $^k\boldsymbol{e}'_{ij}$ are the latent features after $k$ message passing. $l_i$ is the number of the region that $\boldsymbol{v}_i$ belongs to. $f_{\theta_V}$, $f_{\theta_E}$, and $f_{\theta_R}$ are the node, edge, and region feature update functions, respectively. 
\par
\textit{Decoder}: In the decoder, the latent features are transformed back into the original space as follows;
\begin{align}
        \begin{split}
        \boldsymbol{v}^{new}_i &= f_{\varphi_V}^{-1}(^{k}\boldsymbol{v}'_i) + \boldsymbol{v}_i\\
        \boldsymbol{e}^{new}_{ij} &= f_{\varphi_E}^{-1}(^{k}\boldsymbol{e}'_{ij}) + \boldsymbol{e}_{ij}
        \end{split}
\end{align}
$\boldsymbol{v}^{new}_i$, $\boldsymbol{e}^{new}_{ij}$ from $f_{feddback}$ and $f_{forward}$ are utilized to construct the corrected $\hat{G}_t$ and $G_{t+1}$, respectively. 
\par
During manipulation, $\boldsymbol{p}_i$ in topological graph $G_t$ is updated by the keypoint tracking first. Then, the corrective model is applied to refine the whole $G_t$ due to historical graph sequences. Finally, new topological graphs $G_{t+1}$ under actions $\boldsymbol{a}_t$ are calculated by the predictive model.

\subsection{Feedback Control of Features} 
We consider the clothing folding task as a series of lifting-placing processes. The nodes of graphs are moved to desired positions in one operation. 
When folding sleeves, grasp points are set on cuffs to fold the sleeve nodes inwards.
When the sleeve is too long, new grasp points are assigned to the crease on the sleeve for another folding. 
Then, when folding the body, the desired crease is one that minimizes the fabric diameter after folding. 
The vertex nodes are employed as the feature nodes and their expected trajectories are circles about the crease. 
The farthest node from the crease is chosen as the grasping node.
When the maximum diameter of the clothing is smaller than the threshold, the folding task finishes.
\par
During one pick-and-place operation, the expected trajectory is implemented in $4$ stages: approaching, lifting, placing, and leaving. 
In approaching stage, the end-effector moves from the default position to the grasping node and grabs it. 
The default position is where the robot does not obstruct the clothing. 
Afterwards, the robot lifts the grasping node to the highest position on the expected trajectory and forms the clothing into an L-shape in side view. This ensures that the clothing crease moves to the desired position. In the placing stage, all feature nodes are placed to the desired position along a straight trajectory, while the end-effector releases and returns to its default position in leaving stage.
\par
\begin{figure}[!t]\centering
        \includegraphics[width=0.35\textwidth,trim=5 1 5 1,clip]{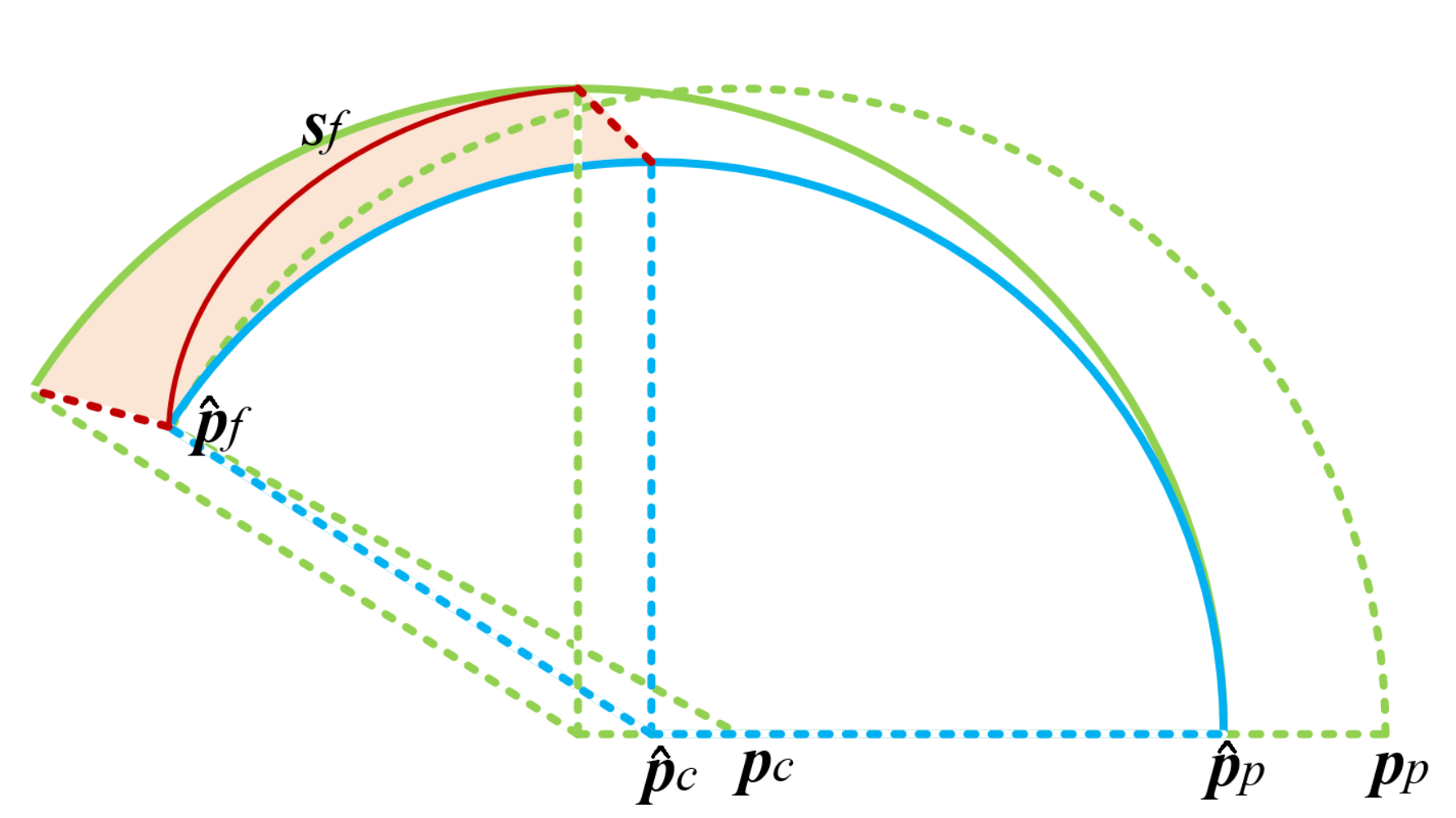}
        \caption{The refined trajectory in lifting stage. The dashed green arc is the initial trajectory. The solid green arc is the moved trajectory. The blue arc is the arc crossing $\hat{\boldsymbol{p}}_f$ and $\hat{\boldsymbol{p}}_p$. The red arc is the interpolated refined trajectory.}
        \label{refine_traj}
\end{figure}

Moreover, to overcome the impact of fabric sliding, we refine the expected trajectory based on feature updates. 
We denote the center of the initial trajectory as $\boldsymbol{p}_c$, the initial placement position of the farthest node as $\boldsymbol{p}_p$, the farthest node, and its placement position in current state as $\hat{\boldsymbol{p}}_f$, $\hat{\boldsymbol{p}}_p$. 
In lifting stage, the interpolation between the arc line crossing $\hat{\boldsymbol{p}}_f$ and $\hat{\boldsymbol{p}}_p$ and the moved initial trajectory is used as the new trajectory. 
The new trajectory, $\boldsymbol{s}_f(\tau), 0\le\tau<1$, is shown in Fig. \ref{refine_traj} and formulated as follows:
\begin{align}
        \begin{split}
                \boldsymbol{s}_f(\tau) &= 
                        \displaystyle\frac{\tau-\tau_0}{1-\tau_0}(R({\boldsymbol{p}}_p,{\boldsymbol{p}}_c,\frac{\pi}{2} (2-\tau))+ \hat{\boldsymbol{p}}_p - \boldsymbol{p}_p)+\\ 
                        &\displaystyle\frac{1-\tau}{1-\tau_0}R(\hat{\boldsymbol{p}}_p,\hat{\boldsymbol{p}}_c,\frac{\pi}{2} (2-\tau))  \\
                \tau_0 &= \frac{2}{\pi} \text{arccos}\frac{(\hat{\boldsymbol{p}}_f -\boldsymbol{p}_c )\cdot(\boldsymbol{p}_c - \boldsymbol{p}_p)}{||\hat{\boldsymbol{p}}_f -\boldsymbol{p}_c || ||\boldsymbol{p}_c - \boldsymbol{p}_p||}\\
        \end{split}
\end{align}
where $\hat{\boldsymbol{p}}_c$ is the center of the new arc. $R(\boldsymbol{p}_1,\boldsymbol{p}_2,\theta)$ is the position where $\boldsymbol{p}_1$ rotates $\theta$ around $\boldsymbol{p}_2$. 
In placing stage, the straight trajectory forwards the new placement position is employed as the new trajectory of each feature node. Taking $\hat{\boldsymbol{p}}_f$ as an example, the refined trajectory, $\boldsymbol{s}_{p_f}(\tau), 1\le\tau<2$, is shown as:
\begin{align}
        \begin{split}
                \boldsymbol{s}_{p_f}(\tau) &=  \frac{\tau- \tau_1}{2 - \tau_1}\hat{\boldsymbol{p}}_p + \frac{2 - \tau}{2 - \tau_1}\hat{\boldsymbol{p}}_f \\
                \tau_1 & = 2 - \frac{||\hat{\boldsymbol{p}}_f  - \hat{\boldsymbol{p}}_p || }{||{\boldsymbol{p}}_c  - {\boldsymbol{p}}_p || }
        \end{split}
\end{align}
\par
By the back propagation of the predictive model, we can obtain the deformation Jacobi matrix $\boldsymbol{J}_f$ between feature nodes and grasping nodes. Then, the robot controller will be:
\begin{align}
        \boldsymbol{v}_r = k_v(\boldsymbol{J}_f^T\boldsymbol{J}_f)^\dagger \boldsymbol{J}_f^T \frac{d}{d\tau}\boldsymbol{s}_f(\tau)
\end{align}
where $k_v$ is a positive velocity coefficient.

\section{Experiments} 
\subsection{Experiment Setup}
The experimental platform is shown in Fig. \ref{pipline} (a). The manipulator, Franka Research 3, is positioned on the plane. Clothing is picked and placed by the Franka hand. Since the gripper is not the focus of our research, we assume that the gripper can stably grasp all layers of clothing for manipulation. 
To construct the topological graph and track its keypoints, a camera is fixed above the platform. 
The algorithm was running on a laptop with Ubuntu 20.04 LTS and an NVIDIA GeForce RTX 4060 Laptop GPU. The manipulator is controlled at a frequency of $30Hz$, synchronized with the camera frame rate. Due to limitations in the speed of Cotracker, feature feedback is provided at $10Hz$.
\par
A jacket-type garment is chosen to assess our method. 
With an opening in the front, the clothing body of Jackets deforms as a folded, single-layer fabric, while the shoulders obstruct its ability to unfold into a flat sheet. 
Its long, unconstrained sleeves can move freely and cause obstruction, further increasing the difficulty.
The jackets can demonstrate the challenges posed by complex structures and are used to verify the effectiveness of our algorithm.
\par
Before manipulation, we grab the shoulders of the jacket and fling it onto the plane. 
This simple action can initially unfold the clothing and expose its overall structure \cite{haFlingBotUnreasonableEffectiveness2022}. However, the movement of the sleeves and hem remains difficult to fully control, and they may obstruct each other. 
Our algorithm starts analyzing from this state and completes both the perception and folding of the garment.

\subsection{Experiments for Topological graph Construction}
\begin{figure}[!t]\centering
        \includegraphics[width=0.48\textwidth,trim=5 1 5 15,clip]{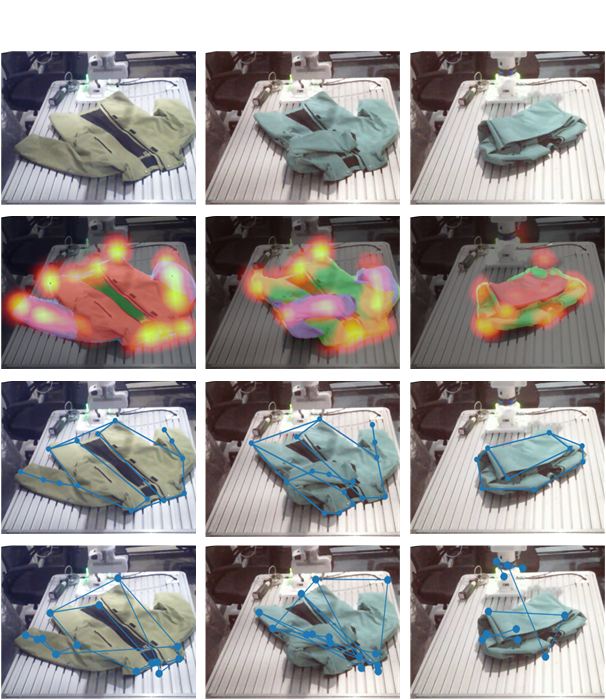}
        \caption{The results of perception and topological graph generation. The first row: the original images. The second row: results of keypoints detection and semantic segmentation. The third row: generated topological graph. The fourth row: results of the traditional clothing keypoints detection.}
        \label{expr1}
\end{figure}

To train the keypoint detection and the semantic segmentation networks, we assembled a dataset comprising 280 images of jackets in various folding configurations. After manual annotation, the dataset was split into a 7:3 ratio for training and validation. The feature recognition and topological graph generation results for clothing in different states are shown in the second and third rows of Fig. \ref{expr1}.
To evaluate the performance of our method, we compare our topological graph extraction method with the traditional semantic keypoint detection approach, which predefines the clothing's topological structure and attempts to match all topological nodes from the image. Both methodologies use the same training dataset and identical neural network architecture. The topological graphs generated by the traditional semantic keypoint detection are shown in the fourth row of Fig. \ref{expr1}.
\par
In Fig. \ref{expr1}, the semantic keypoint detection approach is influenced by the symmetric structure of clothing, making it difficult to distinguish symmetrical sleeves (as shown in the first column). After deformation, challenges in detecting certain keypoints can lead to errors in the generated results (as shown in the second column). Additionally, if the self-occlusion of clothing obscures keypoints, the predefined clothing template struggles to match the actual garment (as the third column), failing to generate a connected topological graph. In contrast, our method successfully extracts the corresponding topological structures from clothing in various folding states. Even when some keypoints are not detected (as shown in the second column), our method can establish correct nodal connections and generate accurate topological graphs.



 
\subsection{Experiments for  Clothing Folding} 

Considering that the topological dynamics is solely dependent on the structure of the clothing, models trained in simulation are expected to exhibit similar performance in real-world applications. Therefore, the GNN was trained using a garment folding dataset \cite{xuegarmenttrackingcategorylevelgarment2023} collected in PyBullet environment. Each folding step in the dataset is decomposed to construct topological graphs separately, thereby enabling the learning of a generalized dynamics model for various topological structures.
To enhance the accuracy of this model, we directly extract the topological graphs from the clothing meshes. Specifically, we first retrieve a primarily decomposed garment by separating the sleeves and stitching in the standard clothing space, thus initializing the initial graph. Subsequently, by comparing the projection of the garment on the plane before and after folding, new creases are generated for each step to update the topological graph.
\par
All mappings in GNNs are performed using MLPs with two hidden layers and a hidden size of 32. The embedding vectors utilized have a dimensionality of 16. The mean square error of the predicted model on the validation dataset is $0.017$, demonstrating the effectiveness of our dynamics model in predicting garment deformation. Moreover, the effectiveness of the dynamics model further validates that our topological representation can effectively capture the motion constraints of clothing during the folding process.
\par
\begin{figure}[!t]\centering
        \includegraphics[width=0.48\textwidth,trim=5 1 5 1,clip]{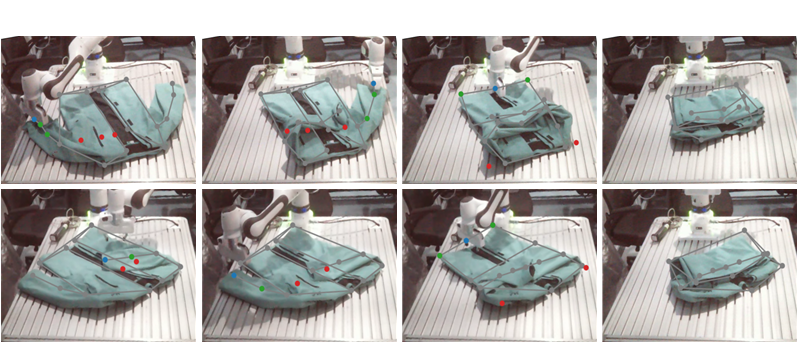}
        \caption{The snapshots of our clothing folding methods. Four sequential phases of two examples are shown from left to right. The topological graph is represented in gray, while the selected grasp nodes are marked in blue, feature nodes in green, and the target nodes in red.}
        \label{expr2}
\end{figure}

To validate the feasibility of our topological dynamics based cloth folding framework, we conduct the experiments to fold jackets from different configurations. Two examples, one with the jacket fully unfolded and the other with the jacket occluded by its sleeves, are shown in Fig. \ref{expr2}.
\par
In Fig. \ref{expr2} the corresponding topological graph, the selected grasp and feature nodes, and their aim nodes are depicted in gray, blue, green, and red, respectively.
In these folding tasks, our framework effectively recognizes the topological structure of complex clothing.  Given the constraints of the operational space, the robotic arm cannot directly reach some aim nodes. In such cases, we lift the grasping node upwards before executing the expected trajectory to move the entire garment. As shown in the final folding step, our model successfully predicts the overall slipping of the clothing and updates the folding trajectory accordingly. Our approach implements the lifting-placing operations, even when the cloth slips during manipulation, ultimately achieving the desired fold.



\section{Conclusion}
In this work, we present a novel topological representation of clothing and the corresponding dynamics model for folding complex garments with wrinkles and self-occlusion.
Experimental results demonstrate that our topological graphs can be extracted from the clothing under various occlusion conditions and effectively capture the constraints associated with folding. The corresponding model is capable of predicting clothing motion during manipulation with high accuracy. 
By integrating the topological graph extraction and dynamic prediction into our Jacobi-based feedback controller, we validate the effectiveness of our method in folding jackets into specified configurations in experimental settings. In the future, it is valuable to extend topology based clothing manipulation methods to other tasks and types of clothing.

\addtolength{\textheight}{-12cm}   








\bibliographystyle{Transactions-Bibliography/IEEEtran}
\bibliography{Transactions-Bibliography/IEEEabrv, Transactions-Bibliography/BIB_xx-TIE-xxxx}

\end{document}